\documentclass{article}

\usepackage{arxiv}

\usepackage[utf8]{inputenc} % allow utf-8 input
\usepackage[T1]{fontenc}    % use 8-bit T1 fonts
\usepackage{hyperref}       % hyperlinks
\usepackage{url}            % simple URL typesetting
\usepackage{booktabs}       % professional-quality tables
\usepackage{amsfonts}       % blackboard math symbols
\usepackage{nicefrac}       % compact symbols for 1/2, etc.
\usepackage{microtype}      % microtypography
\usepackage{lipsum}		% Can be removed after putting your text content
\usepackage{graphicx}
\usepackage{natbib}
\usepackage{doi}
\usepackage{amssymb}            % Defines common symbols like \mathbb R
\usepackage{mathtools}          % Extends amsmath, providing common math tools
\usepackage{mathrsfs}           % Enables \mathscr, which can work in cases that \mathcal does not
\mathtoolsset{showonlyrefs}     % Only number equations that are referenced (optional)
\usepackage{graphicx}           % For including images
\usepackage{subcaption}         % Allows for the use of subfigures and subcaptions
\usepackage[space]{grffile}     % For spaces in image names
\usepackage{url}                % For displaying urls
\usepackage{wrapfig}
\usepackage{algorithm2e}[rule]
\RestyleAlgo{ruled}
\SetKwComment{Comment}{/* }{ */}

\title{Multi-agent Reinforcement Learning for Dynamic Dispatching in Material Handling Systems}

\author{
    Xian Yeow Lee  \\
    xian.lee@hal.hitachi.com \\
    Industrial A.I. Lab.\\
    Hitachi America Ltd.
    \And
    Haiyan Wang \\
    haiyan.wang@hal.hitachi.com \\
    Industrial A.I. Lab.\\
    Hitachi America Ltd.
    \And
    Daisuke Katsumata  \\
    daisuke.katsumata@hal.hitachi.com \\
    JR Automation Collaboration Project \\
    Hitachi America Ltd.
    \And
    Takaharu Matsui \\
    takaharu.matsui@hal.hitachi.com \\
    JR Automation Collaboration Project\\
    Hitachi America Ltd.
    \And
    Chetan Gupta  \\
    chetan.gupta@hal.hitachi.com \\
    Industrial A.I. Lab.\\
    Hitachi America Ltd.
    }

\date{}

\hypersetup{
pdftitle={A template for the arxiv style},
pdfsubject={q-bio.NC, q-bio.QM},
pdfauthor={David S.~Hippocampus, Elias D.~Striatum},
pdfkeywords={First keyword, Second keyword, More},
}

\begin{document}
\maketitle

\begin{abstract}
This paper proposes a multi-agent reinforcement learning (MARL) approach to learn dynamic dispatching strategies, which is crucial for optimizing throughput in material handling systems across diverse industries. To benchmark our method, we developed a material handling environment that reflects the complexities of an actual system, such as various activities at different locations, physical constraints, and inherent uncertainties. To enhance exploration during learning, we propose a method to integrate domain knowledge in the form of existing dynamic dispatching heuristics. Our experimental results show that our method can outperform heuristics by up to 7.4\% in terms of median throughput. Additionally, we analyze the effect of different architectures on MARL performance when training multiple agents with different functions. We also demonstrate that the MARL agents' performance can be further improved by using the first iteration of MARL agents as heuristics to train a second iteration of MARL agents. This work demonstrates the potential of applying MARL to learn effective dynamic dispatching strategies that may be deployed in real-world systems to improve business outcomes. 
\end{abstract}

\section{Introduction}
\label{sec:introduction}

Material handling systems are integral to warehousing and logistics operations across industries, playing a pivotal role in ensuring efficient material flow~\citep{bhandari2023material}. Achieving optimal performance metrics, such as maximizing throughput, within these systems can have cascading effects on downstream business processes, resulting in streamlined operations and improved efficiency. Dynamic dispatching that involves real-time task allocation and resource management is crucial to achieving optimal performance. Traditionally, heuristics dispatching rules, such as shortest route, and nearest location, are usually employed. However, these rules are often sub-optimal in complex material handling systems due to various real-world challenges, such as inherent uncertainties at the input, output, and system dynamics, interconnected sub-processes that leads to complex interactions between the sub-processes, and system changes (such as change in layout or cycle times) due to business expansion or reduction. 

Reinforcement Learning (RL) offers a promising avenue for overcoming the challenges and enhancing dynamic dispatching, allowing algorithms to adapt and optimize decisions in real-time scenarios~\citep{kayhan2023reinforcement, shyalika2020reinforcement}. However, training RL algorithms requires a simulator to mimic real-world complexities for algorithm development and testing, as it is often cost-prohibitive and infeasible to train RL algorithms in actual systems. In this work, we aim to develop a framework to train event-based multi-agent RL (MARL) strategies to improve the key performance index (KPI)s of the material handling systems and contribute to the following: We :\textbf{1)} formulate the dynamic dispatching aspect of material handling systems as an optimization problem, which may be solved using RL, \textbf{2)}: develop an environment that mimics the characteristics and reflects the complexity of an actual conveyor material handling system, \textbf{3)} adapt a MARL algorithm to handle asynchronous multi-agent event-based dynamic dispatching via Monte-Carlo roll-outs, \textbf{4)} apply a method that leverages domain knowledge in the form of heuristics to improve the exploration of the MARL training, \textbf{5)} propose a technique which improves the MARL agents' performance using previous iterations of MARL agents.

\section{Dynamic Dispatching for Material Handling Systems}

\begin{wrapfigure}{r}{0.5\textwidth}
  \centering
  \includegraphics[width=0.5\textwidth]{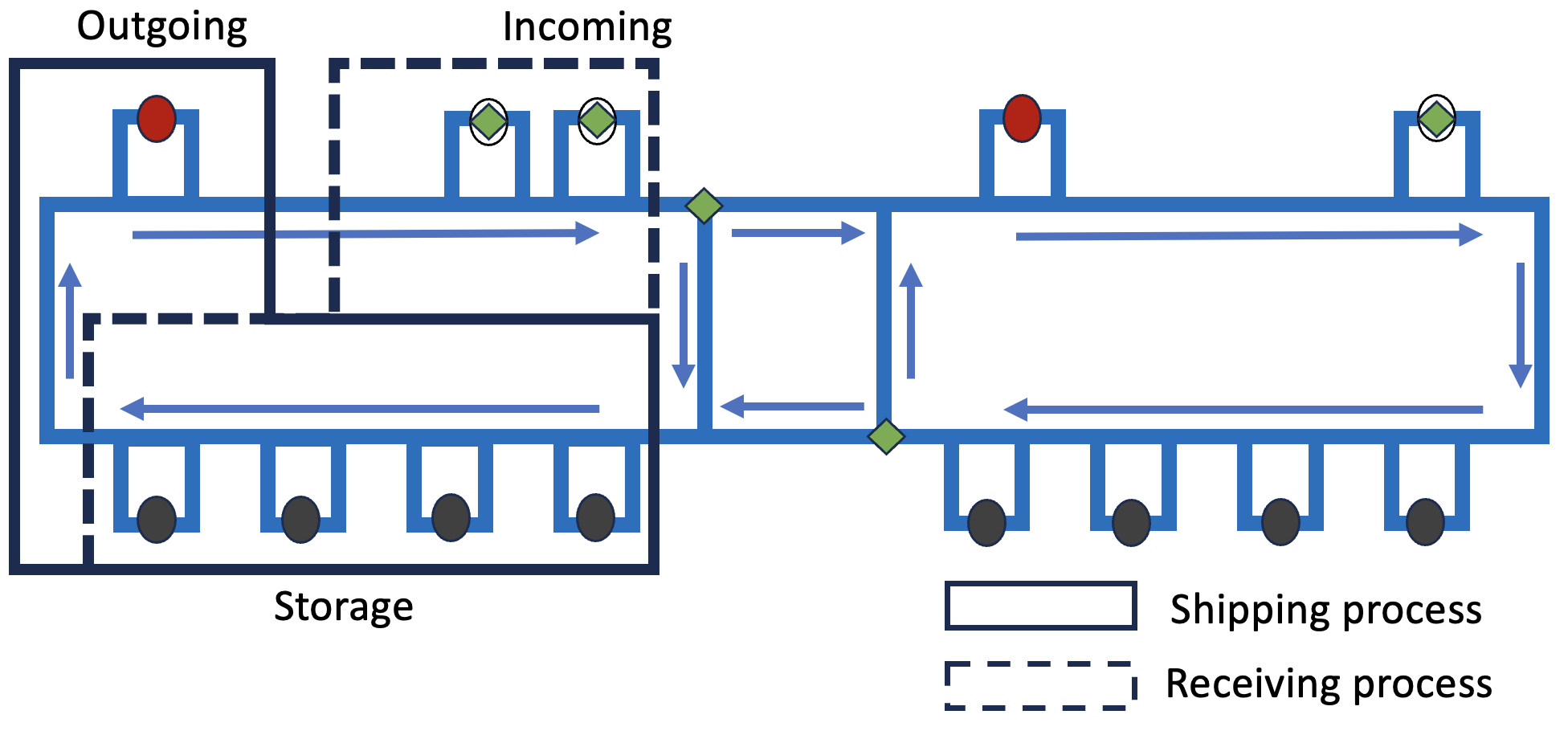}
  \caption{Example layout of a simplified material handling system.}
  \label{fig:simplified_example}
\end{wrapfigure}

In this section, we introduce an instance of a generic material handling system that is widely used in a variety of industries and would benefit from optimized dynamic dispatching strategies. We consider a system that consists of a conveyor belt that transports goods from multiple points to multiple destinations. This system consists of three types of points (incoming points, storage points, and outgoing points) and two major processes (receiving and shipping processes). As an example, we refer readers to a simplified example in Fig.~\ref{fig:simplified_example}. As shown in the figure, material handling systems typically have input points through which the goods enter the system. We represent these as incoming points, illustrated by white circles. At these incoming points, goods are loaded onto pallets which are transported via the conveyor to one of the many storage points, represented by gray circles. The transportation of the goods from the input points to the storage points constitutes the receiving process. Simultaneously, there is a shipping process, which consists of transporting selected goods from the storage points to the outgoing points on pallets. The selection of the goods from the storage to the outgoing points is stochastic and based on the demand of the shipping process. In this system, there is a fixed number of pallets that are constantly circulating in the system. As such, this results in pallets that either carry goods to the storage or outgoing points or empty pallets that are required at the incoming or storage points for loading the goods. 

We consider two main classes of decisions that have to be made in these systems, illustrated by green diamonds in Fig.~\ref{fig:simplified_example}. The first class of decisions are the dispatching decisions made at incoming points, where the system decides which storage points to send the incoming goods to. The second class of decisions represents the dispatching decision made at junctions, where the system decides which direction to send \textit{empty} pallets, which will subsequently affect the number of empty pallets available to feed the incoming and storage points. Thus, the goal of the dynamic dispatching problem here is to make decisions at each of the decision points such that the total receiving throughput (number of pallets entering the storage points) and total shipping throughput (number of pallets entering the outgoing points) are maximized. The main challenge of this problem emerges from the fact that all three points (incoming, storage, and outgoing) are interconnected via a shared and limited resource: the pallets. Furthermore, the complexity of the problem is often exacerbated by constraints imposed by the system's physical design, such as a limit on the number of pallets allowed on certain sections of the conveyor. Additionally, while pallets with goods are routed from their source to destination via the shortest route in the ideal case, sub-optimal decisions could cause the shortest route to be congested, making a longer route more efficient. Thus, an optimal dynamic dispatching strategy must coordinate between different decision points to achieve a high overall throughput.

\subsection{Problem Formulation}
In this section, we first formalize the dynamic dispatching problem as an optimization problem, then posit it as a MARL problem. Based on the description above, the dynamic dispatching problem for material handling can be posed as:

\begin{equation}
\begin{aligned}
\text{Max:} \quad & \sum^{T}_{t=0} F(\boldsymbol{\bar{x}_t}, \boldsymbol{\bar{y}_t}) \\
\text{subject to:} \quad & C_{x_t}(\boldsymbol{\bar{x}_t}) \leq 0, \\
& C_{y_t}(\boldsymbol{\bar{y}_t}) \leq 0, \\
& C_{x_t}, C_{y_t} = H(\boldsymbol{\bar{x}_{t-1}}, \boldsymbol{\bar{y}_{t-1}})
\end{aligned}
\end{equation}

where $F(.)$ denotes the objective function, $\bar x_t, \bar y_t$ denote the optimization variables (in this case, representing the two class of decisions), $C_{x_t}(.)$, $C_{y_t}(.)$ the constraints on $\bar x_t, \bar y_t$, where $C_{x_t}(.)$, $C_{y_t}(.)$ are a consequence of a previous decision $\boldsymbol{\bar x_{t-1}},\boldsymbol{\bar y_{t-1}}$ and $H(.)$ denote any arbitrary transformation. In the context of our application, the objective function is represented by the total throughput, and the constraints on the dispatching decisions are represented in the form of invalid dispatching decisions, which are a function of the previous dispatching decisions (e.g., previous decisions causes certain conveyor sections to be congested, causing the pallets to be re-routed). As can be seen, the dynamic dispatching problem can be viewed as optimizing a sequence of decision variables over a finite horizon in order to maximize the cumulative objective function, which brings us to a typical RL formulation.

Based on the formulation above, we can then pose the optimization problem as a MARL problem under the centralized training decentralized execution (CTDE) paradigm, where the agents are jointly trained using shared information during training, but act independently during deployment and the goal of the MARL algorithm is to maximize the cumulative discounted rewards of the environment. For additional background on MARL and CTDE, we refer readers to Appendix~\ref{appendix:marl_background}. The CTDE MARL approach to the dynamic dispatching problem above has several benefits: 1) Decomposing the joint action space into multiple smaller action spaces allows us to circumvent the curse of dimensionality of a combinatorically large action space when training a single centralized agent, 2) Since the cardinality of the action space of each decision point can potentially be different, representing each decision point as a separate agent allows for greater flexibility and also enables us to handle discrete asynchronous events easier (discussed in further detail below) and 3) As material handling systems often contains replicated layouts, deployment in decentralized fashion potentially allows new decision making points to be added to the environment using the same trained agent with minimal or no additional retraining, thus enabling scalability.

\section{Related works and discussions}
\label{sec:related works}

We briefly discuss existing related works from multiple perspectives: 1) works that are based on heuristic/optimization approaches for dynamic dispatching, 2) works that leverage (MA)RL for dynamic dispatching, and 3) research areas related to ideas of this work. Traditionally, dynamic dispatching has depended mainly on manually but expertly designed rules~\citep{rajendran1999comparative, dhurasevic2018survey, yoon2021dynamic}. These works are not necessarily limited to material handling systems and are generalizable to many industries, but they often require the expertise of a subject matter expert, which has become an increasing challenge due to labor shortages in many industries. Beyond manually-designed rules, there are also efforts to develop optimization-based methods~\citep{jia2017dynamic, gohareh2022simulation, wang2023data}. These methods often employ some version of evolutionary or swarm-based optimization, coupled with simulations, to generate dynamic dispatching policies. Despite the challenges of applying RL to dynamic dispatching~\citep{khorasgani2020challenges}, the research community has  strive to develop RL-based approaches due to RL's potential to generalize and handle uncertainties. We refer interested readers to the following papers for a detailed review of RL-based approaches for dynamic dispatching across various applications~\cite{kayhan2023reinforcement, shyalika2020reinforcement, panzer2021deep, bahrpeyma2022review}. Last but not least, the proposed idea in this paper of using heuristics alongside the MARL policies is just one way to aid exploration during training. We highlight that similar ideas have been explored in works which leverages expert demonstrations~\citep{ramirez2022model} and there are also numerous heuristic-agnostic methods that focus on improving exploration, such as the methods discussed in the following works by~\citet{ladosz2022exploration, yang2021exploration, hao2023exploration}.

\section{Environment}

To evaluate the feasibility of a MARL-based dynamic dispatching approach, we developed a Python-based simulator that serves as a training platform. We benchmarked the simulator by implementing several heuristics and validated that the simulated KPIs reflect the KPIs of an actual proprietary material handling system, thus ensuring the simulator has sufficiently high fidelity. We then develop a training environment following the convention of PettingZoo~\citep{terry2021pettingzoo}.

In our experiments, we consider a three-loop material handling system with a conveyor belt that transports material from the incoming points to the storage and from the storage to the outgoing points, as described above. The system consists of four incoming points, twenty storage points, and six outgoing points, with 500 available pallets. The demand at the outgoing points is modeled according to the statistics of the actual system and is significantly non-uniform. Furthermore, we imposed additional rules on the environment to reflect actual constraints due to the design of the material handling system. Specifically, each incoming, storage, and outgoing point has a designated buffer for the number of pallets that can be in the queue to be processed. If the point's buffer is full, then pallets would be rerouted around the conveyor belt until the buffer is available. Additionally, there is a limit on the maximum number of pallets that can be present on the conveyor belt section connecting the different loops. If the number of pallets is exceeded, the junction points that control the flow of pallets between two different loops may either 1) stop the flow of pallets if both downstream segments of the conveyor are full, thus causing potential congestion upstream or 2) send the pallets on a path that is not the shortest, resulting in a sub-optimal traveling time. Together, these constraints further contribute to the complexity of the dynamic dispatching problem, as optimal dispatching decisions may occasionally be overridden. We refer readers to Appendix Table~\ref{tab:system} for additional details on the specifications of the material handling system we used in our experiments. 

\begin{table}[h]
    \small
    \centering
    \begin{tabular}{l | c}
        \hline
        \textbf{State space} & \textbf{Dimensionality} \\
        \hline
        Process Identifier  &  1 \\
        Number of pallets heading to each storage & 20 \\
        Number of pallets at each junction going into each downstream direction & 4 \\
        Difference between number of outgoing and incoming pallets at each storage & 20 \\
        \hline
        \textbf{Action space} & \textbf{Dimensionality} \\
        \hline
        Receiving agent (decides the storage point) & 20 \\ 
        Junction agent (decides the downstream direction) & 2 \\
        \hline
        \textbf{Reward function} & \textbf{Dimensionality} \\
        \hline
        Total throughput  & 1 \\ 
        \hline
    \end{tabular}
    \caption{State, action and reward definitions of the material handling environment}
    \label{tab:states}
\end{table}

In the environment, we define the state space as shown in Table~\ref{tab:states}. This information was chosen as states to reflect the data that are available in real-time to existing heuristic methods for a fair comparison without giving the RL agents access to additional information. The only additional information we introduced to the state space is the process identifier that allows a centralized agent to distinguish between the different agents at different decision points. Additionally, the environment also returns an event indicator, $I_t$, at every time step, which denotes if an event has occurred and a dispatching decision is required. However, this is not included in the state space of the agents and is only used as information to mask out decisions that are not required at certain dispatching locations due to asynchronous decision making (explained in more detail in the next section). For the action space, we define the actions of each agent as the number of dispatching decisions it can make. In our experiments, we consider two types of agents: agents at the incoming points, which have an action dimension of twenty, and agents at the junctions of the conveyor system, which has an action dimension of two. We define the reward of the environment to be the total throughput (sum of throughput at all storage and outgoing points) across the duration of one hour of operation, which we also define as the length of an episode. While there are other potential metrics that can also be optimized, such as minimizing the idling time or balancing the load across all incoming points, we focus only on total throughput as a metric since it is a broadly applicable metric. However, the approach we proposed in this paper is agnostic to the choice of metric and a different metric could be used in place of the total throughput. In this work, we also did not attempt meticulous state space and reward engineering although it is critical to training successful RL agents, as we wanted to demonstrate the feasibility of a MARL approach using only limited information and a rudimentary form of reward. Finally, to avoid potential instability during training, we normalize all observations to a range of $[0,1]$ and scaled rewards by a factor of 0.01 based on the magnitude of the throughput.

\section{Methods}

In this section, we propose a framework to train MARL-based dynamic dispatching policies that outperform several manually designed heuristics. We conjectured and empirically observed that training vanilla MARL policies would be challenging due to a large combinatorial space of decisions, asynchronous event-based decision-making, and environmental constraints that may override the dispatching decision. On the other hand, businesses and domain experts who operate material handling systems often have invested time into developing heuristics based on their experiences from operating the system. While these heuristics may not be optimal, they are often better than the performance of training a MARL-based policy from scratch. As such, we propose to leverage these heuristics to inject domain knowledge into the learning process of the MARL training. Specifically, during training, we systematically interleave the heuristic's actions with the actions of the MARL policies and store all the actions and associated rewards in memory. Consequently, the MARL policies are trained using transition tuples consisting of its actions and associated rewards and heuristic actions and their associated rewards. By interleaving heuristics' actions into the training process, we utilize the existing domain knowledge as an exploration tool that could potentially guide the MARL policies into a regions of high-performing policies. To concretize our framework, we used a Monte-Carlo version of multi-agent proximal policy optimization (PPO)~\citep{yu2022surprising} with decentralized actors and a centralized critic to train the dynamic dispatching policies. Note that the choice of using PPO in the multi-agent setting is mainly driven by the simplicity of implementation, as compared to more advanced methods that requires the instantiation of multiple classes of networks such as Q-mix~\citep{rashid2020monotonic}, and the observed empirical effectiveness of PPO over more simple value-based methods such as IQL~\cite{tan1993multi}. We refer readers to more discussions on the choice of the MARL algorithm in Appendix~\ref{sec:appendix_RL_choice}. To interleave the heuristic's decisions with the MARL policies' actions, we use a simple switching policy that alternates between the MARL and the heuristic's actions. 

Another unique aspect of this dynamic dispatching scenario is the asynchronous property of the problem. Since the dispatching is event-based (based on the arrival of the pallets at dispatching points) and the occurrence of these events largely depends on the interaction between the previous dispatch decisions and uncertainty in demand at outgoing points, this creates a unique MARL setting where not all agents would make a decision at the same time points. To address this issue, we mask the agents' actions that are not needed and only store the \textit{<state, action, reward, next state>} tuple for a particular agent when it has a dispatching event. %The asynchronous event-based nature of the system also leads to the number of actions taken by each agent being different at the end of each episode. Naturally, this leads to different trajectory lengths across agents and episodes. To handle different trajectory lengths across different episodes when updating the actor's weights in PPO's clipped surrogate objective~\citep{schulman2017proximal}, we find that truncating the trajectories' lengths to the shorter trajectory length works well in practice (see Appendix~\ref{appendix:ppo} for more details).
The pseudo-code of the asynchronous event-based Monte-Carlo multi-agent PPO is shown in Algorithm~\ref{alg:mappo}.

\begin{algorithm}
\small
\caption{Asynchronous event-based multi-agent PPO}\label{alg:mappo}
\textbf{Initialize:} Environment, Num. Episodes $N$, Episode Horizon $T$\\
\textbf{Initialize:} Actors $\pi_{\theta_1}$, $\dots$, $\pi_{\theta_n}$, Critic(s) $V_{\phi}$, Heuristic function $H(.)$\\
\While{Episode $\leq N$}{
    \While{t $\leq T$}{
        \For{every Actor $\pi_{\theta}$}{
            Observe state $s_t$ and event indicator $I_t$ from environment \\
            \uIf{$t$ is even \textbf{and} $I_t$ is True}{
                $a_t = \pi_{\theta}(s_t)$ \Comment*[r]{Use actor's actions}
            }
            \uElseIf{$t$ is odd \textbf{and} $I_t$ is True}{
                $a_t = H(s_t)$ \Comment*[r]{Use heuristic's actions}
            }
            \uElse{Skip $a_t$  \Comment*[r]{Non-event transition}}
            Observe reward $r_t$ and next state $s_{t+1}$ \\
            Store $(s_t, a_t, r_t, s_{t+1})$ in memory if $I_t$ is True
        }
    } 
    % \textbf{Truncate trajectories:}\\
    % \For{each Actor $\pi_{\theta}$}{
    %     \textbf{Get current trajectory length $L_{\text{current}}$} \\
    %     \textbf{Get previous trajectory length $L_{\text{previous}}$} \\
    %     $L_{\text{min}} = \min(L_{\text{current}}, L_{\text{previous}})$ \\
    %     Truncate the current trajectory to length $L_{\text{min}}$ \Comment*[r]{Truncate trajectories}
    % }
    
    \textbf{Update actors' weights $\theta_{i}$:}\\
    \For{each actor $i$}{
        \[
        L^{CLIP}(\theta_i) = \mathbb{E}_t \left[ \min \left( r_t(\theta_i) \hat{A}_t, \text{clip}(r_t(\theta_i), 1 - \epsilon, 1 + \epsilon) \hat{A}_t \right) \right]
        \]
        Perform gradient ascent on $L^{CLIP}(\theta_i)$ \Comment*[r]{Update using PPO's clipped surrogate objective}
    }

    \textbf{Update critic's weights $\phi$:}\\
    \[
    L^{V}(\phi) = \mathbb{E}_t \left[ \left( V_{\phi}(s_t) - R_t \right)^2 \right]
    \]
    Perform gradient descent on $L^{V}(\phi)$ \Comment*[r]{Update using PPO's critic objective}
}
\end{algorithm}

\section{Experiments and Results}

\subsection{Comparing MARL-based policies with existing heuristics}
For our initial experiments, we consider a simpler version of the dynamic dispatching problem, where we only optimize the decisions at the incoming points to dispatch pallets to storage points rather than jointly optimizing the decisions at the incoming and junction points. For comparison, we implemented three heuristics that were designed manually by experts. The first heuristic sends the pallets to random storage points in the same loop. This is based on the principle that storage points within the same loop are closer in distance to storage points in other loops. The second heuristic we consider is a set of rules that select the optimal storage point based on the number of pallets in each storage point's buffer, the distance to each storage point, and the storage point's number of incoming and outgoing pallets. Finally, the third heuristic we implemented is similar to the second one, with an additional rule that considers the congestion at the junctions between the loops. The algorithms on the three heuristics developed are presented in Appendix~\ref{sec:appendix_heuristics}. We denote the first, second, and third heuristics as 'Low', 'Medium' and 'High' based on their respective evaluated performance. To ensure fair benchmarking, we had systematically fine-tune the hyperparameters associated with each heuristic (if any) to ensure that the performance exhibited by each heuristic is the maximized. For the dispatching decisions at the junctions, we implemented a heuristic that sends empty pallets into the direction of the conveyor loops, which have the least number of pallets. We used these three heuristics in our framework to train three instances of MARL-based dynamic dispatching policies. In each experiment, the material handling system consists of four incoming points, resulting in four trainable policies. We trained each policy for 300 episodes and saved the best policy with the best evaluation rewards during training. All training were performed on a 12-core 3.5 GHz Intel Core i9 CPU. For evaluation, we compared the total throughput averaged across 150 episodes initialized with unique random seeds. 

\begin{figure}[h]
    \centering
    \includegraphics[width=0.8\textwidth]{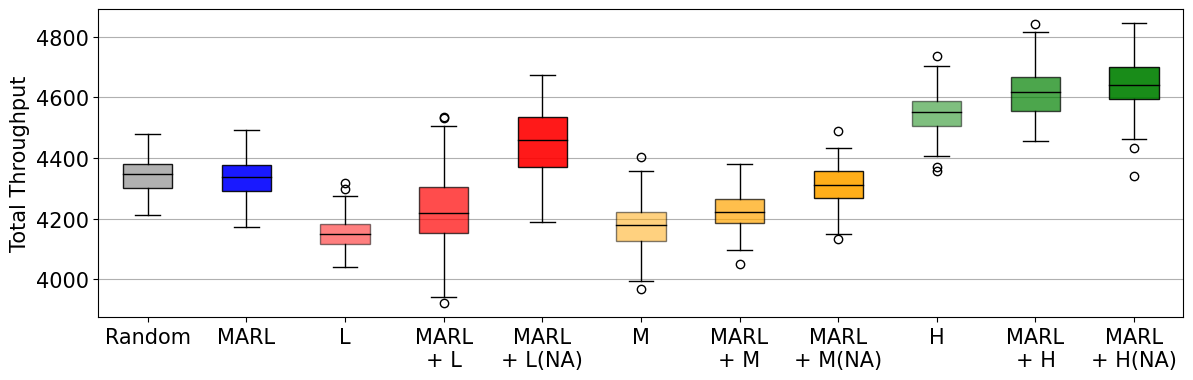}
    \caption{Distributions of throughput during evaluation of different strategies. 'L', 'M', 'H' denotes manually designed heuristics, 'MARL' denotes a vanilla training procedure, 'MARL + X' denote MARL policies trained with heuristics and legends with '(NA)' denote evaluation without heuristics.}
    \label{fig:barplot}
\end{figure}

Fig.~\ref{fig:barplot} shows the throughput performance of each heuristic, compared with their MARL counterpart during evaluation. In the figure, 'Random' denotes a baseline policy where the dispatch decisions are random among the storage points across loops. 'L', 'M', 'H' denotes the 'Low', 'Medium', and 'High' heuristics respectively. 'MARL' denotes a vanilla multi-agent PPO training procedure, while 'MARL + X' denotes the proposed method where the MARL's decisions are interleaved with heuristic's decisions during the training. Furthermore, during evaluation, we have two options: We could interleave the heuristic decisions with MARL decisions, similar to the training setting, or we could switch off the heuristic's decision and depend entirely on the MARL's dispatching decision. We denote the latter option as 'MARL + X (NA)' in the figure, representing non-assisted evaluation.

Based on the results, we drew several conclusions. First, we observed that the 'Random' policy forms a relatively strong baseline and outperforms the 'Low' and 'Medium' heuristics, respectively. We hypothesize that this is due to the inherent stochasticity of the system that even a manually designed heuristic using knowledge of the system may sometimes under-perform a random policy. Next, we see that training a multi-agent PPO in a vanilla way does not outperform a 'Random' heuristic. This further demonstrates the complexity of environment as a naive MARL-implementation is only on-par with the random policy. Nevertheless, in all three experiments that integrate heuristics in the MARL training process, the total throughput of MARL policies outperforms the total throughput of pure heuristics. More interestingly, the non-assisted version of dispatching outperforms the dispatching that interleaves with the heuristic during evaluation. We highlight that this phenomenon is simultaneously counter-intuitive and intuitive. On one hand, we had expected the performance of the MARL policy without heuristic assistance (MARL + X(NA)) to deteriorate since there is potentially a shift in the distribution of the state visitations between training and testing settings due to the presence of the heuristic taking alternate actions during training and the absence of it during testing. Nevertheless, we also hypothesized that interleaving the heuristic's decision during evaluation could limit the true potential of the MARL's policies, especially if the heuristics are sub-optimal. Based on the results, we infer that the policies learned by the MARL algorithm are robust enough to the distribution change between training and evaluation that the presence of the heuristic decisions is more detrimental than beneficial. In Table~\ref{tab:improvements}, we tabulate the improvement of the median throughput of each instance of MARL policy over their respective heuristics' median throughput. We observed a monotonic decrease in terms of improvement for the non-assisted throughput as the performance of the heuristic gets better. Hence, the more sub-optimal a heuristic, the larger the potential for marginal gains when training the MARL policies with the heuristic. This observation leads to another point with practical impact: a heuristic with a decent performance is not necessarily needed or assumed. Even if we begin with a sub-optimal heuristic like the 'L' heuristic, our proposed method can still improve the total throughput over the existing heuristic, hence creating a net positive business impact, although starting better heuristics can ultimately lead to better results. 

\begin{table}{} % Adjust the width as needed \begin{wraptable}
    \small
    \centering
    \begin{tabular}{l | c | c}
        \hline
        \textbf{Method} & \textbf{\% Improvement (Assisted)} & \textbf{\% Improvement (Non-Assisted)} \\
        \hline
        RL + L & 1.68 & 7.44 \\
        RL + M & 0.98 & 3.13 \\
        RL + H & 1.43 & 1.98 \\
        \hline
    \end{tabular}
    \caption{Percentage improvement of MARL approaches trained with different heuristics compared with the respective heuristic's original performance.}
    \label{tab:improvements}
\end{table}

\subsection{Comparing different architectures for MARL-policies with different functions}

\begin{figure}[h]
    \centering
    \includegraphics[width=0.8\textwidth]{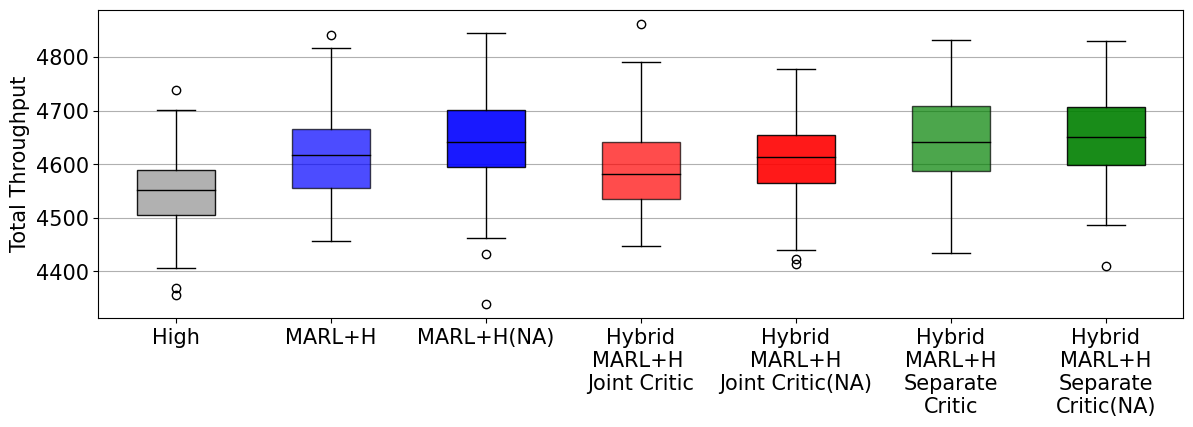}
    \caption{Distributions of throughput during the evaluation of a heuristic strategy, compared with MARL with a joint critic v.s. separate critics when training policies with different action spaces.}
    \label{fig:barplot_junc}
\end{figure}

Having validated that MARL policies trained with heuristics can outperform a pure heuristic, we experimented with replacing both classes of dispatching decisions with MARL agents, specifically one class of agents that dispatches at the incoming points and one class that dispatches at the junctions between loops. In this scenario, we trained eight agents, four at the incoming points and four at the junction points. While the state space of these two types of agents is the same, the action spaces are different. Since the functions of both types of agents are different, we explore two different architectures for training these MARL policies using our proposed framework. In the first architecture, we explored a joint critic architecture, where a single critic network is used for estimating the values of all actors. In the second architecture, we consider a separate critic architecture, where we have one critic network for the actors at the incoming points and another for the actors at the junction points. Since we only implemented one heuristic for the junctions, we used that heuristic's decision to interleave with the actors' decisions at the junction during training. Fig.~\ref{fig:barplot_junc} illustrates the results of these experiments. For comparison, we also visualized the performance of the 'High' heuristic and the best MARL result from Fig.~\ref{fig:barplot}. We denote the experiments that use MARL for both the incoming and junction points as 'Hybrid-MARL', and use the same convention where 'NA' represents the evaluation setting that does not use the heuristic's decisions. Based on the distributions of throughput in Fig.~\ref{fig:barplot_junc}, we observed that, in general, training multiple types of MARL policies does not pose any additional significant challenges, as they all converged to the same ballpark of performance during evaluation. Nonetheless, we note that using a joint critic leads to a lower performance as opposed to using separate critics. These results implied that sharing information between the two classes of agents during training via a joint critic wouldn't improve performance but further deteriorate the learning process. Similar to before, we observed that using MARL policies without the heuristics during evaluation results in a similar or slightly better performance. 

\subsection{Decoupling MARL policies from influences of heuristics}
\begin{figure}[h]
    \centering
    \includegraphics[width=0.8\textwidth]{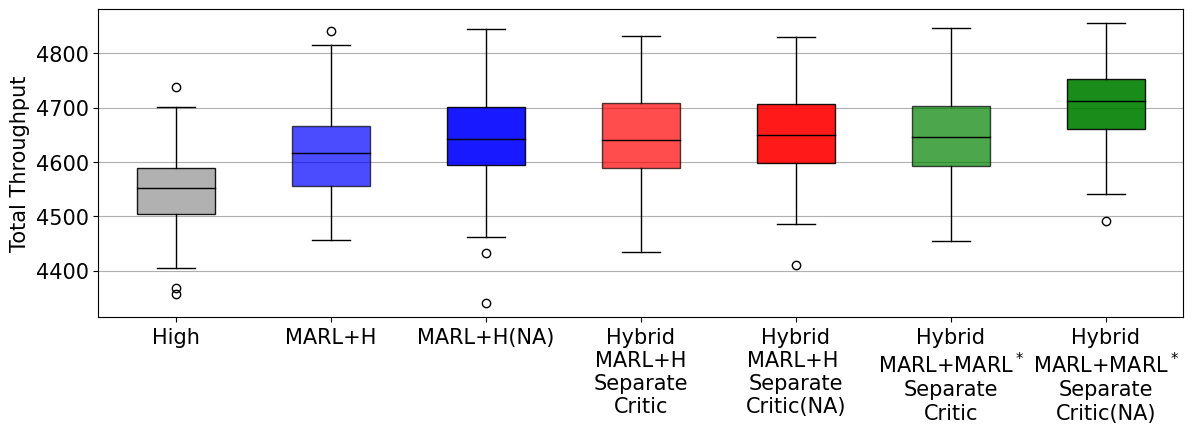}
    \caption{Distributions of throughput during evaluation of a heuristic strategy, MARL policies trained heuristics and MARL policies trained with previous iterations of MARL policies.}
    \label{fig:barplot_sp}
\end{figure}

Motivated by the observation that the presence of the heuristic potentially limits the MARL's performance during evaluation, we conducted another set of experiments where we trained a second iteration of MARL policies from scratch. However, in this iteration, we use the first iteration of MARL policies as the heuristics by freezing the weights of the policies rather than the original set of heuristics. Our intuition is that since the first iteration of MARL agents performs better without heuristics during evaluation, they act as better exploration tools than the original set of heuristics. We performed this set of experiments using the same experimental parameters as before. Fig.~\ref{fig:barplot_sp} illustrates the evaluation results, where we show the progression from the best heuristic to the application of MARL policies at the incoming points, then to the application of MARL policies at both receiving and junction points and finally to the second iteration of MARL policies that were trained with previous iterations of trained policies, denoted as 'MARL + MARL$^*$'. As shown in the figure, we observed that once again, removing heuristic decisions during evaluation enables a higher total throughput, and training the MARL policies with a previous iteration of MARL policies results in marginally higher throughput. Overall, by using the initial best heuristic to train the first iteration of MARL policies and then repeating the process to train a second iteration of MARL policies, we increased the median throughput from 4552 to 4712, representing a 3.51\% increase in total throughput when compared with the best heuristic, and also increased the median throughput from 4349, representing an 8.34\% improvement when compared to the random dispatching strategy. For detailed statistics of the results and an analysis of the decisions made by the MARL and heuristic policies, please refer to Table~\ref{tab:extra_results} and Fig.~\ref{fig:actions} in the Appendix.

% Relation to transfer learning/curriculum learning
% Relation to exploration

\section{Conclusion}

Dynamic dispatching strategies are critical to increasing throughput in material handling systems, which are widely used in many industries. In this work, we propose an event-based MARL framework to learn an instance of a dynamic dispatching strategy. Using a MARL framework enables us to potentially scale to arbitrary systems sizes. To enhance the performance of the MARL algorithm, we proposed a method to leverage existing domain knowledge in the form of heuristics to improve the exploration capability and to yield a multi-agent strategy that outperforms the best heuristic by 3\% in an environment that reflects the complexity of an actual material handling system. We also showed that the trained MARL policies could be used as fixed heuristics to train a newer set of policies that are independent of the original heuristics. Future work will focus on integrating more sophisticated exploration methods and solving the challenges of deployment in an actual system.

\bibliographystyle{unsrtnat}
\bibliography{references}  %%% Uncomment this line and comment out the ``thebibliography'' section below to use the external .bib file (using bibtex) .

\appendix

\section*{Appendix}
% No label, since this can't be referenced meaningfully with \ref{}
\section{Background on PPO and MARL}

\subsection{Proximal Policy Optimization (PPO)}\label{appendix:ppo}

Reinforcement Learning (RL) is a type of machine learning where an agent learns to make decisions by interacting with an environment. The agent's goal is to learn a policy \(\pi(a|s)\), which maps states \(s\) to actions \(a\), in a way that maximizes the cumulative reward over time. The agent receives feedback from the environment in the form of rewards, which it uses to update its policy. Traditional methods in RL include value-based methods, such as Q-learning, and policy-based methods, such as policy gradient algorithms.

Policy gradient methods directly parameterize the policy \(\pi_\theta(a|s)\) using parameters \(\theta\) and optimize these parameters by maximizing the expected cumulative reward. One of the simplest forms of policy gradient is the REINFORCE algorithm, which updates the policy parameters \(\theta\) in the direction of the gradient of the expected reward:

\[
\nabla_\theta J(\theta) = \mathbb{E}_t \left[ \nabla_\theta \log \pi_\theta(a_t|s_t) \hat{R}_t \right],
\]

where \(\hat{R}_t\) is the return, representing the cumulative reward from time step \(t\). Although effective, traditional policy gradient methods can suffer from high variance and instability due to large updates to the policy.

In this work, we have used the Proximal Policy Optimization (PPO) as the choice of algorithm within our MARL framework.
PPO is a popular algorithm in the field of RL and an improvement over traditional policy gradient methods, designed to optimize the policy in an efficient and stable manner.

PPO operates by iteratively improving a stochastic policy \(\pi_\theta(a|s)\), where \(\theta\) denotes the parameters of the policy network, \(s\) represents the state, and \(a\) represents the action. The key innovation in PPO is the use of a clipped surrogate objective function, which aims to constrain the magnitude of policy updates, thus preventing large, potentially destabilizing changes to the policy. The objective function for PPO can be expressed as:

\[
L^{CLIP}(\theta) = \mathbb{E}_t \left[ \min \left( r_t(\theta) \hat{A}_t, \text{clip}(r_t(\theta), 1 - \epsilon, 1 + \epsilon) \hat{A}_t \right) \right],
\]

where \(r_t(\theta) = \frac{\pi_\theta(a_t|s_t)}{\pi_{\theta_{\text{old}}}(a_t|s_t)}\) is the probability ratio of the new policy to the old policy, \(\hat{A}_t\) is the advantage function estimating the relative value of action \(a_t\) at state \(s_t\), and \(\epsilon\) is a hyperparameter controlling the range of clipping.
%Note that since we have opted for a Monte-Carlo style update (refer to Appendix~\ref{sec:appendix_RL_choice} for more details on the choice of training), the probability ratio essentially takes the form of 
% \[r_t(\theta) = \sum_{t=0}^{T} \frac{\pi_\theta(a_t|s_t)}{\pi_{\theta_{\text{old}}}(a_t|s_t)}\]. 

% \noindent Since there is no guarantee that the current and previous trajectories are of the same length, this necessitates the truncation step shown in Algorithm~\ref{alg:mappo}. 

In our implementation, we have also used an actor-critic architecture where the we trained an additional centralized critic network, parameterized by $\phi$ to estimate the value of a state, $V_{\phi}(s)$. Subsequently, the advantage function, \(\hat{A}(s_t)\), is derived from $V_{\phi}(s_t) - R_t$. Note that here, $R_t$ is the reward from the environment and is different from the probability ratio $r_t(\theta)$. 

The clipping mechanism in the objective function above serves to penalize the new policy if it diverges too much from the old policy, ensuring updates remain within a trusted region. This helps in maintaining the stability and reliability of learning. PPO is widely used due to its effectiveness across various RL tasks and its ability to be implemented with minimal tuning compared to other complex algorithms.

\subsection{Multi-Agent Reinforcement Learning (MARL)}\label{appendix:marl_background}

Multi-Agent Reinforcement Learning (MARL) extends the principles of RL to environments where multiple agents interact and learn concurrently. Each agent in a MARL setting aims to maximize its own cumulative reward, which often depends on the actions of other agents, leading to a dynamic and interactive learning process. In the context of our work, the reward function (the total throughput) is also a shared reward function, hence creating a cooperative scenario. These complexities introduces challenges such as non-stationarity, where the environment's dynamics change as other agents learn and adapt.

In a MARL environment, each agent \(i\) has its own policy \(\pi_{\theta_i}(a_i|s_i)\), where \(\theta_i\) denotes the parameters of agent \(i\)'s policy network, \(a_i\) represents the action taken by agent \(i\), and \(s_i\) represents the state observed by agent \(i\). The goal of each agent is to maximize its own expected return \(J_i(\theta_i)\):

\[
J_i(\theta_i) = \mathbb{E} \left[ \sum_{t=0}^{T} \gamma^t r_i^t \right],
\]

where \(r_i^t\) is the reward received by agent \(i\) at time step \(t\) and \(\gamma\) is the discount factor. 

\textbf{Centralized Training with Decentralized Execution (CTDE)}

A popular paradigm in MARL is Centralized Training with Decentralized Execution (CTDE). This approach leverages the advantages of centralized information during training while allowing agents to operate independently during execution. During the training phase, agents can access the global state and the actions of other agents, facilitating more coordinated and efficient learning. However, during execution, each agent acts based solely on its local observations, ensuring scalability and robustness in decentralized settings.

Mathematically, let \(\mathcal{S}\) denote the global state space and \(\mathcal{O}_i\) denote the observation space of agent \(i\). During training, each agent's policy \(\pi_{\theta_i}\) can condition on the global state \(s \in \mathcal{S}\) and the actions of other agents \(\{a_j\}_{j \neq i}\). The centralized value function \(V(s, \{a_i\}_{i=1}^n)\) can be used to estimate the joint value of the state and actions, leading to better-informed policy updates.

During execution, each agent \(i\) uses its decentralized policy \(\pi_{\theta_i}(a_i | o_i)\), where \(o_i \in \mathcal{O}_i\) is the local observation. This ensures that agents can operate independently and react to their local environment without requiring centralized coordination, which is crucial for real-world applications where communication may be limited or costly. CTDE strikes a balance between leveraging global information to enhance learning and maintaining the practicality of decentralized decision-making, making it a powerful approach in multi-agent systems. In our implementation, due to the asynchronous aspects of the environment, we only condition each policy on the global, shared state $s_t$ and not on the actions of other policies. 

\section{Heuristic's details}\label{sec:appendix_heuristics}

In this section, we detail the three heuristics that were used as baselines in the paper. These heuristics were developed manually developed based on the domain's experts intuition and further fine-tuned using the simulation model to achieve the best results. 

To begin, we define several auxiliary variables and functions that may be used in each of the heuristics respectively. In general, each storage point in the system belongs to one of the many "loops" of the conveyor system. As such, for any pair of storage and incoming point, we can define if the storage point belong to the same loop as the incoming point or a different loop. Hence, we define $\mathbb{S_\text{same}}$ as the set of all storage points that belong to the same loop as the incoming point, $\mathbb{S_\text{other}}$ as the set of all storage points that belong to the other loops, and $\mathbb{S_\text{all}}$ as the set of all storage points, where $\mathbb{S_\text{all}}$ = $\mathbb{S_\text{same}} \cup \mathbb{S_\text{other}}$. Furthermore, we also define $In(s)$ as a function that returns the number of incoming pallets assigned to a given storage point, and $Out(s)$ returns the number of outgoing pallets being retrieved from the storage point from all the outgoing points. Additionally, we also define $X_{same}$ as the number of pallets assigned to all storage points in the same loop for a given incoming point, and $X_{other}$ as the number of pallets assigned to all storage points in the other neighbouring loops for a given incoming point. Finally, in Heuristic 2, we also defined a cost function $minCost(L)$ Specifically, the cost function is defined as:

\begin{equation}
\text{minCost($L_{i}$)} = \frac{X_{\text{same}} - X_{\text{min}}}{X_{\text{max}} - X_{\text{min}}}  + C_{L_{j}} 
\end{equation}
where $X_{max}$ and $X_{min}$ denotes the number of pallets in the loop with the maximum/minimum number of pallets in the entire system and $C_{L_{j}}$ is a constant scalar cost value that is proportional to the distance it takes to send a pallet from loop $i$ to loop $j$. Note that $C_{L_{j}}$ is a parameter that can be tuned by the user. Additionally, the Medium and High heuristics also requires an initialization of several parameters, which can be fine-tuned by the user depending on the system characteristics~\footnote{Specific values used in this paper have been intentionally withheld to protect proprietary information.}.

\begin{algorithm}
\small
\caption{Heuristic 1 (Low)}\label{alg:h1}
\textbf{Initialize:} Environment, Num. Episodes $N$, Episode Horizon $T$\\
\While{Episode $\leq$ N}{
    \While{t $\leq$ T}{
        Observe state $s_t$ and event indicator $I_t$ from environment \\
        \uIf{$I_t$ is True}{
            $\mathbb{S}\leftarrow$ $\mathbb{S_\text{same loop}}$ \Comment*[r]{Get set of storage points within the same loop} 
            $a_t \sim \mathbb{S}$ \Comment*[r]{Sample and dispatch to random storage point}
        }
        \uElse{Skip $a_t$  \Comment*[r]{Non-event transition}
        }
    }
}
\end{algorithm}

\begin{algorithm}
\small
\caption{Heuristic 2 (Medium)}\label{alg:h2}
\textbf{Initialize:} Environment, Num. Episodes $N$, Episode Horizon $T$\\
\textbf{Initialize:} $C_1$ \\
\While{Episode $\leq N$}{
    \While{t $\leq T$}{
        Observe state $s_t$ and event indicator $I_t$ from environment \\
        \uIf{$I_t$ is True}{
            $\mathbb{S} \leftarrow \mathbb{S_\text{all}}$ \Comment*[r]{Get set of all storage points} 
            $\mathbb{S} \leftarrow \{s \in \mathbb{S} \mid \text{In}(s) \leq C_1\}$ \Comment*[r]{Get storage points with fewer than $C_1$ incoming pallets}
            $\mathbb{S} \leftarrow minCost(\mathbb{S})$ \Comment*[r]{Get set of storage points in loop with min. cost}
            \uIf{$|\mathbb{S}| = 1$}{
                $a_t \leftarrow s \in \mathbb{S}$ \Comment*[r]{If only one storage point in set, select it}
            }
            \uElse{
                $a_t \leftarrow \arg\min_{s \in \mathbb{S}} (\text{In}(s))$\Comment*[r]{Select storage with smallest num. of incoming pallets}
            }
        }
        \uElse{
            Skip $a_t$ \Comment*[r]{Non-event transition}
        }
    }
}
\end{algorithm}

\begin{algorithm}
\small
\caption{Heuristic 3 (High)}\label{alg:h3}
\textbf{Initialize:} Environment, Num. Episodes $N$, Episode Horizon $T$\\
\textbf{Initialize:} $C_1$, $C_2$, $C_3$ \\
\While{Episode $\leq N$}{
    \While{t $\leq T$}{
        Observe state $s_t$ and event indicator $I_t$ from environment \\
        \uIf{$I_t$ is True}{
            get $X_{same}$ \Comment*[r]{Get num. of pallets assigned to storage points in same loop} 
            get $X_{other}$ \Comment*[r]{Get num. of pallets assigned to storage points in other loop} 
            \uIf{$X_{same}$ < $C_1$ and $X_{other}$ < $C_2$}{
                $\mathbb{S} \leftarrow \mathbb{S_\text{all}}$ \Comment*[r]{Get set of all storage points} }

            \uElseIf{$X_{same}$ < $C_1$ and $X_{other}$ > $C_2$}{
                $\mathbb{S} \leftarrow \mathbb{S_\text{same}}$ \Comment*[r]{Get set of storage points in same loop} 
            }
            \uElseIf{$X_{same}$ > $C_1$ and $X_{other}$ < $C_2$}{
                $\mathbb{S} \leftarrow \mathbb{S_\text{other}}$ \Comment*[r]{Get set of storage points in other loop} 
            }
            \uElse{
                $\mathbb{S} \leftarrow \mathbb{S_\text{all}}$ \Comment*[r]{Get set of all storage points} 
            }

            $\mathbb{S} \leftarrow \{s \in \mathbb{S} \mid \text{In}(s) \leq C_3\}$ \Comment*[r]{Get storage points with fewer than $C_3$ incoming pallets}

            \uIf{$\mathbb{S} \leftarrow \mathbb{S} \setminus \{S_{others}\}$ $\neq$ $\varnothing$}
                {
                   $\mathbb{S} \leftarrow \mathbb{S} \setminus \{S_{others}\}$ } \Comment*[r]{Remove storage points that belong to other loop}

            \uIf{$|\mathbb{S}| = 1$}{
                $a_t \leftarrow s \in \mathbb{S}$ \Comment*[r]{If only one storage point in set, select it}
            }
            \uElse{
                $a_t \leftarrow \arg\min_{s \in \mathbb{S}} (\text{Out}(s) - \text{In}(s))$\Comment*[r]{Select storage point with min. out-in difference}

            }
        }
        \uElse{
            Skip $a_t$ \Comment*[r]{Non-event transition}
        }
    }
}
\end{algorithm}

\newpage

\section{Additional results and hyperparameter details}

The following tables below tabulates the details on the experimental results shown in this paper. Table~\ref{tab:system} details the specifications of the material handling system simulation we used as the training environment. Note that these specifications are selected to reflect the complexities of an actual material handling systems. Table~\ref{tab:training} lists the training hyper-parameters we used when training the multi-agent PPO algorithm. Note that the training hyper-parameters used to train the multi-agent policies are simply default hyper-parameters, while the hyperparameters used in the heuristics have been tuned and optimized to the system. As such, we believe this demonstrates the efficacy of our method as the results of the MARL approach could still potentially improve based on additional hyperparameter tuning and we plan to explore the effect of more extensive hyperparameter tuning in future works. Lastly, Table~\ref{tab:extra_results} tabulates the statistics of the box-plots shown in the main paper for all the experiments we conducted. 

\begin{figure}[htbp]
    \small
    \centering
    \begin{subfigure}[b]{0.45\textwidth}
        \centering
        \begin{tabular}{l|c}
        \hline
        \textbf{Specification} & \textbf{Value} \\
        \hline
        Number of Loops & 3  \\
        Number of Incoming Points & 4  \\
        Number of Storage Points & 20  \\
        Number of Outgoing Points & 6  \\
        Number of Junction Points & 4  \\
        Incoming Points Processing Time & 5 sec  \\
        Storage Points Processing Time & 10 sec  \\
        Outgoing Points Processing Time & 6 sec  \\
        Junction Points Processing Time & 0.5 sec  \\
        Buffer size for Incoming Points & 4\\
        Buffer size for Storage Points & 8\\
        Buffer size for Outgoing Points & 10\\
        Number of Pallets & 500 \\
        Simulation resolution & 0.1 sec/step \\
        \hline
        \end{tabular}
        \caption{Material Handling System Specifications}
        \label{tab:system}
    \end{subfigure}
    \hfill
    \begin{subfigure}[b]{0.45\textwidth}
        \centering
        \begin{tabular}{l|c}
            \hline
            \textbf{Critic Network} &   \\
            \hline
            Number of dense layers & 3  \\
            Hidden dimension & 64  \\
            Non-linearity & ReLU \\
            \hline
            \textbf{Actor Network} &   \\
            \hline
            Number of dense layers & 3  \\
            Hidden dimension & 64  \\
            Non-linearity & ReLU \\
            \hline
            Optimizer & AdamW  \\
            Learning rate & 0.001 \\
            PPO clipping parameter, $\epsilon$ & 0.2  \\
            Discount rate, $\gamma$ & 0.99 \\
            Number of train episodes & 300 \\
            Simulation steps per episode & 36000  \\
            \hline
        \end{tabular}
        \caption{MARL training hyper-parameters}
        \label{tab:table2}
    \end{subfigure}
    \caption{Additional details on training setup}
    \label{tab:training}
\end{figure}

\begin{table}[h]
    \small
    \centering
    \begin{tabular}{l|c|c|c|c|c}
        \hline
        \textbf{Experiments} & \textbf{Min} &\textbf{1st Q}&\textbf{Median}&\textbf{3rd Q}&\textbf{Max}\\
        \hline
        Random & 4213 & 4301  & 4349 & 4380 & 4479 \\
        Low & 4040 & 4118 & 4150 & 4183 & 4317\\
        Medium & 3970 & 4128 & 4180 & 4221 & 4405 \\
        High & 4357 & 4505 & 4552 & 4589 & 4738 \\
        MARL & 4173 & 4292 & 4338 & 4376 & 4491 \\
        MARL + Low (Assisted) & 3922 & 4154 & 4220 & 4303 & 4534\\
        MARL + Low (Non-Assisted) & 4190 & 4369 & 4459 & 4535 & 4673 \\
        MARL + Medium (Assisted) & 4051 & 4185 & 4221 & 4263 & 4379\\
        MARL + Medium (Non-Assisted) & 4134 & 4269 & 4311 & 4355 & 4489 \\       
        MARL + High (Assisted) & 4456 & 4555 & 4617 & 4665 & 4841\\
        MARL + High (Non-Assisted) & 4340 & 4594 & 4642 & 4700 & 4845\\
        Hybrid MARL + High, Joint Critic (Assisted) & 4447 & 4534 & 4581 & 4641 & \textbf{4862}\\
        Hybrid MARL + High, Joint Critic (Non-Assisted) & 4414 & 4565 & 4614 & 4653 & 4777\\
        Hybrid MARL + High, Separate Critic (Assisted) & 4435 & 4588 & 4641 & 4708 & 4832\\
        Hybrid MARL + High, Separate Critic (Non-Assisted) & 4410 & 4597 & 4650 & 4707 & 4830\\
        Hybrid MARL + MARL$^*$, Separate Critic (Assisted) & 4454 & 4592 & 4645 & 4703 & 4847\\
        Hybrid MARL + MARL$^*$, Separate Critic (Non-Assisted) & \textbf{4491} & \textbf{4660} & \textbf{4712} & \textbf{4752} & 4855\\

        \hline
    \end{tabular}
    \caption{Numerical statistics of box-plots shown in results. Best results are highlighted in bold font.}
    \label{tab:extra_results}
\end{table}

\section{Discussion on other MARL approaches}
\label{sec:appendix_RL_choice}

In this section, we briefly discuss several challenges we faced over the course of the experiments. In addition to multi-agent PPO, we had also set out to try multi-agent value-based methods. We had experimented with simple algorithms such as Independent Q-learning~\citep{tan1993multi}, but find that the policies did not exhibits signs of learning, despite trying out common explorations strategies and also the proposed method of interleaving expert heuristic within the training process. Nonetheless, we plan to revisit this approach with more sophisticated value-based methods in the future such as Value-Decomposition Networks (VDN)~\citep{sunehag2017value} and Q-mix~\citep{rashid2020monotonic}. 

We also experimented with a more conventional version of multi-agent PPO, where the value estimates are bootstrapped mid-episode rather than using a Monte-Carlo style update. We observed from our experiments that this version tend to converge to a similar performance slightly faster than the Monte-Carlo version, however at the cost of greater instability during training. Hence, we have opted for a Monte-Carlo training to keep the workflow simpler. Last but not least, we also performed a simple hyper-parameter tuning in terms of the actor and critic's network architectures but find that it is not a major factor in terms of the final evaluation performance, thus the results presented are based on a commonly used hyper-parameters rather than optimized hyper-parameters. 

\section{Visualization of heuristic v.s. MARL decisions}
\begin{figure}[h]
    \centering
    \includegraphics[width=0.8\textwidth]{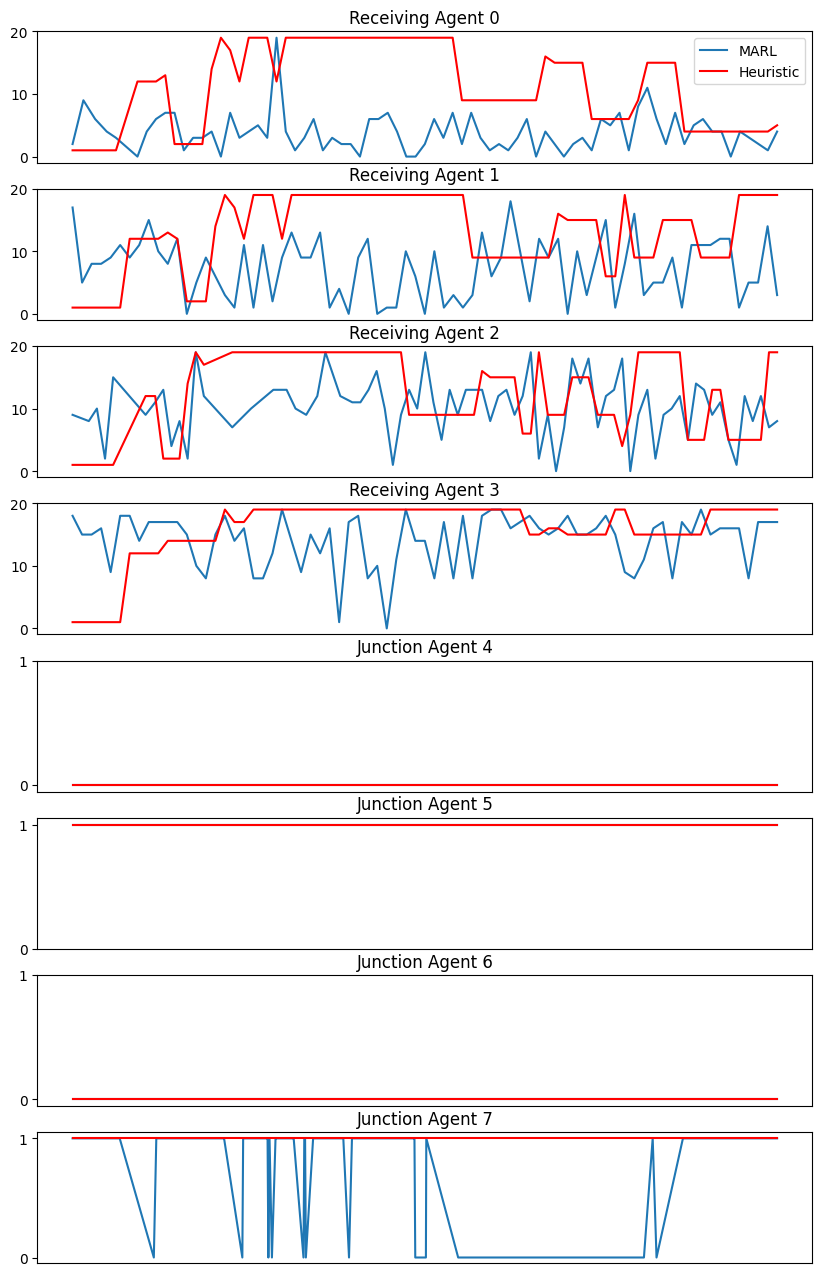}
    \caption{Visualization of actions taken by MARL policies v.s. the best heuristic (H). The total throughput for the heuristic is 4634 and 4854 for the MARL policies in this evaluation.}
    \label{fig:actions}
\end{figure}

In Figure~\ref{fig:actions}, we visualize and compare the decisions made by the best MARL policy (Hybrid MARL + MARL$^*$, Separate Critic and Non-Assisted) with the decisions of the best heuristic (H). An interesting observation is that in the first four subplots, representing the dispatching at the incoming points, we see that the heuristics decisions are relatively stable. In contrast, the decisions of the MARL policies are significantly more erratic and there are almost no agreements between the MARL and heuristic in terms of the decisions made at the same time, except for a few events. This observation implies that a policy that results in a better total throughput would most likely require the design of a very complex heuristic, which may not be feasible to be designed manually. From the bottom four subplots, we observe that surprisingly, the heuristics consistently makes the same dispatching decision. Nevertheless, in three of the subplots, the MARL policies essentially imitates the decisions of the heuristic, while in the last subplot, the MARL policy sometimes deviates from the heuristic. We take this observation as an additional validation that the MARL policies are capable of making a stable, consistent decisions and only make decisions that deviate from the heuristic when it results in a higher reward.

\section{Broad discussions on deployment and generalizations}

In this section, we discuss the broader implications of our proposed framework on various industries and some practical challenges in deploying such a solution. Although we utilized a conveyor system in warehouses as a specific case study to demonstrate the strengths of the proposed method, we envision that the proposed method of combining Multi-Agent Reinforcement Learning (MARL) with existing heuristics to learn a dispatching policy is broadly generalizable to any material handling system that consists of multiple, possibly asynchronous, dynamic decision-making agents. This generalization implies that our method is broadly applicable to numerous industries, such as mining and agriculture, where raw and intermediary materials often need to be transported to several downstream locations. The timing of these transportation activities is often stochastic and dependent on numerous upstream processes. It is worth noting that even a small improvement in efficiency in many of these applications can result in a significant positive impact due to the sheer volume of material being transported.

Regarding deployment challenges, several aspects need to be considered. One main limiting factor or assumption is the existence of a simulator, as it is often impractical to run MARL algorithms in actual scenarios. However, this assumption is not limited to MARL methods but also applies to other optimization-based and heuristic approaches, as domain experts usually validate their methods on a simulator before deploying any heuristics. In terms of computational requirements, the most computationally intensive part of developing the proposed approach is the training phase. We have demonstrated that we can train MARL policies efficiently for a material handling system equivalent in scale to actual systems on conventional consumer hardware. While this is just one example, we do not believe that computational requirements are a limiting factor in deploying such solutions, given recent advancements in computing capabilities.

Two remaining aspects to consider are integration with existing infrastructure and synchronization of the multi-agent policies. Regarding integration, we limited the state space of the MARL agents to information previously used by existing heuristics, so we do not foresee the need for additional infrastructure or sensors for the input. On the output side, while different applications and industries use different frameworks and infrastructures, we note that during deployment, the trained policy is simply a set of numerical weights that can be instantiated in Python, which can be easily wrapped with commercial APIs to communicate with existing infrastructure.

Conventional MARL approaches often assume a synchronous decision-making step among the different agents, which can be unrealistic when deployed. In our work, we explicitly designed the framework to be asynchronous, circumventing the issue of synchronous decision-making. Additionally, as described in the main paper, using the Centralized Training with Decentralized Execution (CTDE) training paradigm allows us to run the multiple agents independently in a decentralized fashion during deployment, further simplifying the task of deployment. Essentially, once the multi-agent RL policies are sufficiently trained together, each policy can be wrapped in an API that communicates with the existing infrastructure to receive sensor information when queried and returns a dispatching decision.

Finally, it is worth noting that in our specific application, we assumed communication latency is not an issue since the frequency of dispatching decisions is several magnitudes slower than the latency to transfer the relatively low-dimensional data streams. However, it is important to consider latency during deployment if the state space requires higher-dimensional data such as image or video input.

\end{document}